\documentclass{article} 
\usepackage{iclr2027_conference,times}

\usepackage{amsmath,amsfonts,bm}

\def\eqref#1{equation~\ref{#1}}

\def\1{\bm{1}}

\DeclareMathAlphabet{\mathsfit}{\encodingdefault}{\sfdefault}{m}{sl}
\SetMathAlphabet{\mathsfit}{bold}{\encodingdefault}{\sfdefault}{bx}{n}

\definecolor{myblue}{rgb}{0.21,0.49,0.74}
\usepackage[colorlinks=true, urlcolor=myblue]{hyperref}
\usepackage{url}
\usepackage{booktabs}
\usepackage{multirow}
\usepackage{graphicx}
\usepackage{wrapfig}
\usepackage[table]{xcolor}
\usepackage{amssymb}
\title{EpiCon: Collective Agent Learning through Co-Evolving Multimodal Memory}

\author{
Ziyun Zeng\textsuperscript{1,2}\thanks{Work done while at MIT-IBM Computing Research Lab.}
\hspace{0.5em}
Hang Hua\textsuperscript{1}\thanks{Corresponding Author.}
\hspace{0.5em}
Shaden Alshammari\textsuperscript{1,3}
\hspace{0.25em}
Rogerio Feris\textsuperscript{1}
\hspace{0.5em}
\\
\hspace{0.03em} \bf William T. Freeman\textsuperscript{3}
\hspace{0.5em}
\bf Jiebo Luo\textsuperscript{2}
\\
\textsuperscript{1} MIT-IBM Computing Research Lab ~~
\textsuperscript{2} University of Rochester ~~ \\
\textsuperscript{3} Massachusetts Institute of Technology\\
{\small \{zzeng24,jluo\}@cs.rochester.edu, hang.hua1@ibm.com, \{shaden,billf\}@mit.edu,
\small rsferis@us.ibm.com} \\
}

\iclrfinalcopy 
\begin{document}

\maketitle
\begin{abstract}
Agents can learn from past executions, but enabling different agents to reuse and build on one another's experience remains challenging. We introduce EpiCon, a shared multimodal memory framework for agent collective learning without updating host model parameters. EpiCon links question-level memory evolution to a persistent experience bank through two independently trained 2B models: a memory controller and a tree self-organizer. The controller jointly refines textual guidance and visual evidence across attempts and selectively includes visual memory. The self-organizer consolidates lessons hierarchically and retrieves experience and rules for new problems. We evaluate EpiCon on eleven benchmarks spanning four multimodal task domains, using two harnesses and multiple backbones. A frozen bank improves other systems even with a single solving attempt. A second harness raises the original system's macro-average score by 2.6 points across eleven benchmarks. Across four host configurations, EpiCon improves macro-average scores by 1.7 to 4.9 points over No Memory and reduces memory-operation time by 67\% to 74\% relative to backbone-sized memory models. Resources available at \href{https://zzzmyyzeng.github.io/EpiCon}{https://zzzmyyzeng.github.io/EpiCon}.
\end{abstract}

\section{Introduction}
\label{sec:intro}

Multi-agent systems (MAS) built on language models combine reasoning, tool use, and collaboration to solve complex tasks~\citep{wu2023autogen,fourney2024magentic,hua2024mmcomposition,zeng2026automated,lin2026lowpowar}. Their executions produce experience about solution procedures, failure modes, and relevant evidence. External memory allows this experience to inform subsequent inference without updating host model parameters~\citep{wang2024agent,ouyang2026reasoningbank}. As tasks accumulate, a system can both draw on earlier experience and contribute new lessons. This motivates a shared memory that supports continued learning within the same MAS while keeping experience useful across different harnesses and backbones.

Recent work has advanced agent memory through linked notes and hierarchical experience structures~\citep{xu2026mem,zhang2026g}, while shared memory systems enable experience reuse across models and frameworks~\citep{tang2025agent,chang2026memcollab}. Multimodal memory preserves visual evidence alongside textual guidance~\citep{zeng2026mementogui}. Lessons may depend on diagram regions or document details, and both guidance and supporting evidence may need revision across attempts. Accumulated lessons also require consolidation for later retrieval and reuse. We study how to connect feedback-driven multimodal refinement with a shared experience bank that different agent systems can reuse and update.

We introduce \textbf{EpiCon} (\textbf{Epi}sodic \textbf{Con}solidation), a shared multimodal memory framework that supports agent collective learning through experience accumulation and reuse. Its dedicated memory harness connects temporary question-level memory with a persistent experience bank. Within a question, a Memory Controller supports textual and visual memory co-evolution, jointly revising actionable guidance and the associated image regions in response to successive attempts and feedback. As the guidance evolves, the controller can revise the visual evidence and adaptively decide whether to include visual memory in the next attempt. Across questions, a Tree Self-Organizer organizes and consolidates lessons in a shared bank maintained independently of the MAS. Accumulated experience can guide later solving within the same MAS and remain useful even if the harness or backbone changes. During bank construction and expansion, different harnesses can both reuse existing experience and contribute new lessons, allowing the evolved bank to support subsequent solving by its contributors. We implement memory control and tree organization with two independently trained 2B models.

We evaluate EpiCon on eleven benchmarks across four multimodal task domains. Experiments with two MAS harnesses and multiple backbones show improved task performance and reuse of experience across configurations. An evolved bank that incorporates experience from another harness also improves the original harness's subsequent solving. The trained 2B models reduce memory-operation time relative to backbone-sized memory models, and performance gains extend to Codex with GPT-5.6-Luna. Our contributions are summarized as follows:
\begin{itemize}
\item \textbf{Collective agent learning through shared multimodal memory.}
We introduce \textbf{EpiCon}, which lets systems share experience without updating host parameters. Frozen-bank reuse improves performance across harnesses and backbones with a single attempt. Contributions from another harness raise the original system's macro-average score by 2.1--3.2 points.

\item \textbf{Co-evolution of textual guidance and visual evidence.}
A memory controller jointly refines guidance and supporting image regions across attempts and selectively includes visual memory. A tree self-organizer consolidates lessons hierarchically and retrieves experience and rules for new problems.

\item \textbf{Efficient memory management with compact models.}
Both modules are independently trained 2B models. On eleven benchmarks spanning four multimodal domains, the 2B variant improves macro-average scores by 1.7--4.9 points over No Memory across four host configurations and reduces memory-operation time by 67--74\% relative to backbone-sized memory models.
\end{itemize}

\section{Related Work}
\subsection{Multi-Agent Systems}
Large language model (LLM) agents are increasingly organized into multi-agent systems (MAS) to divide complex tasks across specialized roles and coordinate complementary capabilities. Existing systems support collaboration through role-based dialogue and structured workflows~\citep{li2023camel,wu2023autogen,chen2024agentverse,hong2024metagpt,qian2024chatdev}, multi-agent debate~\citep{du2023improving}, and adaptive interaction structures such as optimized agent graphs, orchestration, and team generation~\citep{zhuge2024language,fourney2024magentic,yuan2025evoagent}. These approaches make communication and coordination central to MAS design. However, long-horizon and repeated interactions introduce a further requirement: agents must retain useful observations, decisions, failures, and collaboration patterns beyond the current exchange. Communication determines how information moves among agents, whereas memory determines what remains available over time. Recent work therefore begins to model multi-agent memory explicitly, including agent-specific and cross-trial collaboration histories~\citep{zhang2026g}. Our work treats memory as a persistent substrate through which multimodal experiences from different agents can be organized and reused.
\begin{figure*}[t]
\vspace{-2mm}
\centering
\includegraphics[width=\textwidth]{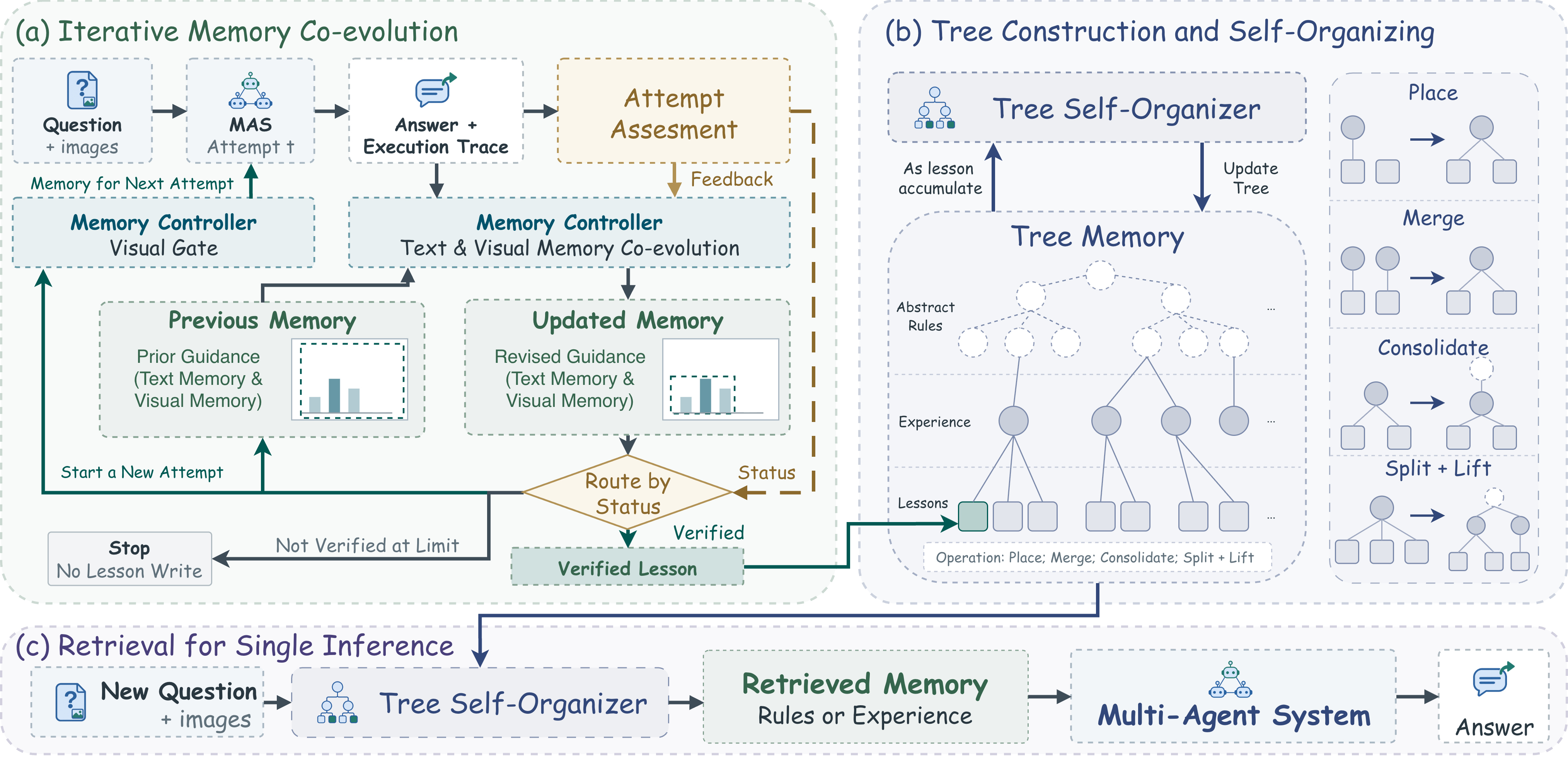}
\vspace{-6mm}
\caption{Overview of EpiCon. The memory controller refines question-level memory, while the tree self-organizer maintains accumulated lessons for subsequent retrieval and reuse. The memory harness coordinates these operations with MAS execution. Bank maintenance occurs during construction; the historical bank remains frozen during evaluation.}
\label{fig:epicon_framework}
\vspace{-5mm}
\end{figure*}
\subsection{Agent Memory and Multimodal Experience}
Agent memory has evolved from retaining information to managing reusable experience. Early systems store and retrieve interaction histories or episodic records to support persistent behavior~\citep{park2023generative,packer2023memgpt,zhong2024memorybank}. Later work adds extraction, consolidation, hierarchical organization, associative retrieval, dynamic linking, and updating~\citep{chhikara2025mem0,xu2026mem,gutierrez2024hipporag}. Other approaches store trajectories, workflows, or learned skills as reusable experience~\citep{zheng2024synapse,wang2023voyager,wang2024agent,zeng2026mira,hua2025v2xum}. Most language-agent memories remain primarily textual. Multimodal agents preserve perceptual evidence alongside higher-level knowledge for long-horizon reasoning~\citep{li2024optimus,long2026seeing,zeng2026mementogui,hua2024finematch,hua2025finecaption}. Beyond reasoning, multimodal systems have also been developed for visual generation\citep{yu2024promptfix,yu2025omnipaint,yu2026aurora}, together with benchmarks and agentic evaluators for visual generation \citep{hua2025mmigbench,zeng2026videoargus}. In MAS, agents may contribute different parts of the same experience, making memory sharing important~\citep{zhang2026g}. EpiCon jointly refines textual guidance and visual evidence across attempts and shares the resulting experience across systems.
\subsection{Reflection and Self-Evolving Agents}
Reflection and self-improvement study how agents can turn interaction outcomes into better future behavior. Self-refinement and critique methods revise outputs using model-generated or external feedback~\citep{madaan2023self,shinn2023reflexion,gou2024critic}, while reflection- and search-based agents use execution outcomes to guide later attempts~\citep{shinn2023reflexion,zhou2023language}. Experiential-learning methods go further by abstracting trajectories into reusable lessons, workflows, policies, or skills~\citep{zheng2024synapse,wang2023voyager,wang2024agent,li2025vquala,zhao2024expel,zhang2024agent,hu2023promptcap,zheng2025skillweaver}. Recent work on self-evolving agents extends this process by integrating successful and failed experience into persistent reasoning memory or broader capability updates~\citep{ouyang2026reasoningbank,yan2026openskill,yang2026towards,cheng2026mem2evolve}. These works form a progression from retry, to reflection, to experience consolidation. Our work connects multimodal memory evolution within a question with the reuse of accumulated experience across questions. We study these two roles separately, evaluating historical experience reuse with a single solving attempt so that its benefits do not depend on additional attempts.
\section{Data Curation}
\label{sec:data}

We construct two supervision datasets for memory control and tree organization using Qwen3.8-27B~\citep{qwen38} as the initial teacher and Qwen3.8-Flash-Next~\citep{qwen3.8flashnext} for selective refinement. The initial teacher proposes joint text--visual memory updates, lessons, and tree-operation demonstrations. Qwen3.8-Flash-Next revises unsuccessful updates, improves lesson quality, and corrects tree-operation targets where needed.

\textbf{Source datasets.}
We collect experience from tasks in MathNet~\citep{alshammari2026mathnet}, MathV360K~\citep{shi2024math}, ChartQA~\citep{masry2022chartqa}, InfoVQA~\citep{mathew2022infographicvqa}, DocVQA~\citep{mathew2021docvqa}, and ChartNet~\citep{kondic2026chartnet}. Our training tasks cover three domains: visual mathematics, document understanding, and reasoning over charts and infographics. MAS executions on these tasks provide the records we use to construct memory updates, lessons, and tree-operation supervision.

\textbf{Memory controller.}
\label{sec:data:controller}
We collect MAS trajectories using CAMEL~\citep{li2023camel}, Codex~\citep{openai2026codex}, and DeepSeek-Harness~\citep{deepseek-harness2026}. Given the question, images, previous memory, latest attempt, and correctness feedback, the teachers generate and refine joint text--visual memory updates. We replay the previous and proposed memories with the same MAS host, retaining Repair transformations that turn failed attempts into successful ones and Compress transformations that preserve success while shortening text without increasing crop area. Reference answers are used to assess attempts during this offline screening. Accepted joint updates provide approximately 25K supervision examples.

\textbf{Tree self-organizer.}
\label{sec:data:tree}
Lessons derived from the collected controller memories seed small memory banks, with historical lessons separated from retrieval queries. The teachers generate and selectively refine demonstrations of placement, merging, splitting, consolidation, and retrieval. We retain approximately 6.5K operation demonstrations that pass schema, node-reference, partition, and provenance checks, including valid empty retrievals when no candidate applies. These checks establish structural validity; individual tree operations are not validated through downstream replay. Appendix~\ref{sec:appendix:data} provides construction and filtering details.

\section{Methodology}
\label{sec:method}
\begin{figure*}[t]
\vspace{-2mm}
\centering
\includegraphics[width=\textwidth]{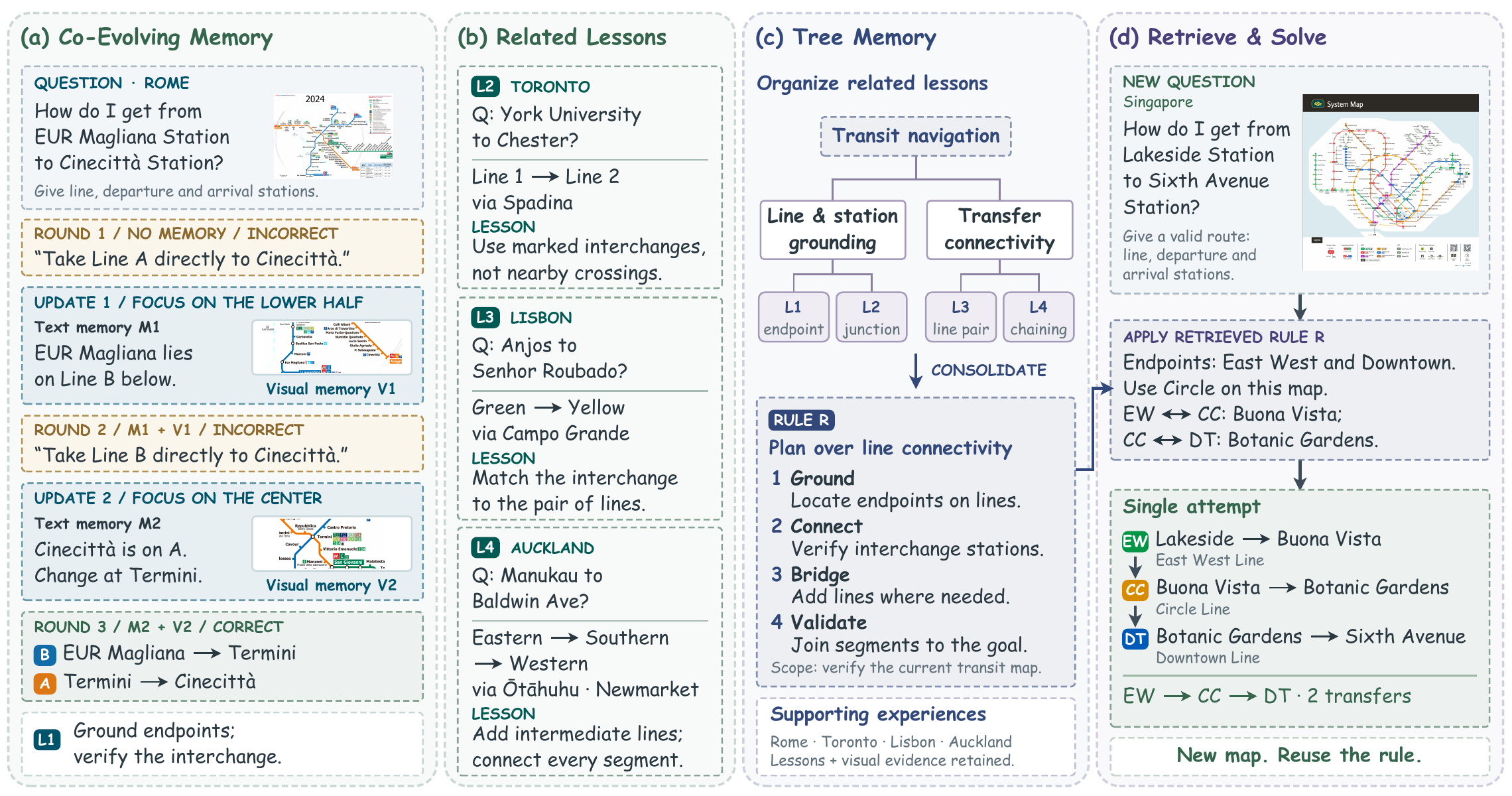}
\vspace{-4mm}
\caption{Illustrative example of EpiCon. (a) Textual and visual memory co-evolve across attempts. (b) and (c) Related lessons are organized and consolidated into a reusable routing rule. (d) Retrieved guidance supports a single solving attempt on a new map. Trajectories and tree operations are reconstructed for illustration.}
\label{fig:metro_memory_showcase}
\vspace{-4mm}
\end{figure*}
EpiCon combines a dedicated memory harness with two independently trained 2B models: a Memory Controller $C_\theta$ and a Tree Self-Organizer $S_\phi$ (Figure~\ref{fig:epicon_framework}). The controller updates textual and visual memory within a question, while the organizer maintains a shared experience bank for reuse across questions and agent systems. The harness coordinates memory updates, retrieval, visual cropping, and storage. Both models are trained on the supervision described in Section~\ref{sec:data}; the host MAS retains its own orchestration and backbone, whose parameters remain unchanged.

\subsection{Shared Multimodal Memory}
\label{sec:method:setup}

EpiCon maintains a persistent experience bank $B$ independently of the MAS that uses it. A system can retrieve guidance from this bank and contribute lessons from its own task executions. These lessons support subsequent tasks within the same MAS and can also be reused across different harnesses and backbones. Collective learning therefore takes place through the accumulation and reuse of shared experience, without updating host model parameters.

For a question $x_i$ with its images and public task information, the organizer retrieves experience $R_i$ once from $B$. The MAS uses the retrieved experience to produce an answer $\hat y_{i,t}$ together with an execution trace $\tau_{i,t}$. When question-level evolution is enabled, the controller maintains a separate memory $W_{i,t}$, initially empty, and revises it using the question, images, existing memory, answer, and execution trace. Our evaluations of historical experience reuse disable this loop and use a single solving attempt.

During bank construction or expansion, accepted lessons and their revisions are incorporated into $B$, allowing experience to accumulate across tasks. Let $\mathcal U_k$ denote the batch of accepted lessons and revisions processed at maintenance step $k$:
\begin{equation}
B^{(k+1)}=S_\phi(B^{(k)},\mathcal U_k).
\label{eq:epicon-bank-update}
\end{equation}
The index $k$ counts maintenance steps, not individual lessons or questions, and a step can process multiple lessons together. The update represents the organizer's proposals after validation and execution by the memory harness. During evaluation, the shared bank is frozen and updates remain local to the current question.

\begin{table*}[t]
\vspace{-4mm}
\centering
\caption{Main results across two MAS harnesses, two backbones, and eleven benchmarks. All methods make a single solving attempt per question, without question-level memory updates. Both EpiCon variants use MC + Tree. Parentheses identify the memory models: ours 2B uses two independently trained 2B models; named backbones operate zero-shot. Time and token usage separate MAS execution from memory operations.}

\label{tab:main_results}
\begingroup
\setlength{\aboverulesep}{0pt}
\setlength{\belowrulesep}{0pt}
\setlength{\cmidrulesep}{0pt}
\setlength{\minrowclearance}{1.5pt}
\providecommand{\mainrowshade}{4}
\definecolor{domainCost}{HTML}{493F7B}
\definecolor{domainDoc}{HTML}{35664D}
\definecolor{domainCode}{HTML}{005A53}
\definecolor{domainMath}{HTML}{00536B}
\definecolor{domainVL}{HTML}{32487A}
\newcolumntype{D}{>{\cellcolor{domainDoc!\mainrowshade}}c}
\newcolumntype{V}{>{\cellcolor{domainCode!\mainrowshade}}c}
\newcolumntype{M}{>{\cellcolor{domainMath!\mainrowshade}}c}
\newcolumntype{R}{>{\cellcolor{domainVL!\mainrowshade}}c}
\newcolumntype{C}{>{\cellcolor{domainCost!\mainrowshade}}c}
\setlength{\tabcolsep}{3.4pt}
\renewcommand{\arraystretch}{1.10}
\newcommand{\benchmarkheader}[1]{%
  \rotatebox[origin=lb]{60}{\strut\textbf{\scriptsize #1}}}
\resizebox{\textwidth}{!}{%
\begin{tabular}{lll DDD VVV MM RRR CC CC}
\toprule
\shortstack[c]{\textbf{MAS}\\\textbf{Harness}} &
\shortstack[c]{\textbf{MAS}\\\textbf{Backbone}} &
\textbf{Memory System} &
\multicolumn{3}{c}{\cellcolor{domainDoc!9}\color{domainDoc}\shortstack[c]{\textbf{Document}\\\textbf{Understanding}}} &
\multicolumn{3}{c}{\cellcolor{domainCode!9}\color{domainCode}\textbf{Visual-to-Code}} &
\multicolumn{2}{c}{\cellcolor{domainMath!9}\color{domainMath}\shortstack[c]{\textbf{Vision-Grounded}\\\textbf{Math}}} &
\multicolumn{3}{c}{\cellcolor{domainVL!9}\color{domainVL}\shortstack[c]{\textbf{General VL}\\\textbf{Reasoning}}} &
\multicolumn{2}{c}{\cellcolor{domainCost!9}\color{domainCost}\textbf{Time ($\times$)}} &
\multicolumn{2}{c}{\cellcolor{domainCost!9}\color{domainCost}\textbf{Tokens ($\times$)}} \\
& & &
\cellcolor{domainDoc!9}\color{domainDoc}\benchmarkheader{DocVQA2026} &
\cellcolor{domainDoc!9}\color{domainDoc}\benchmarkheader{MP-DocVQA} &
\cellcolor{domainDoc!9}\color{domainDoc}\benchmarkheader{ParseBench} &
\cellcolor{domainCode!9}\color{domainCode}\benchmarkheader{Vision2Code} &
\cellcolor{domainCode!9}\color{domainCode}\benchmarkheader{Omni-I2C} &
\cellcolor{domainCode!9}\color{domainCode}\benchmarkheader{ChartMimic} &
\cellcolor{domainMath!9}\color{domainMath}\benchmarkheader{MATH-Vision} &
\cellcolor{domainMath!9}\color{domainMath}\benchmarkheader{WeMath} &
\cellcolor{domainVL!9}\color{domainVL}\benchmarkheader{WorldBench} &
\cellcolor{domainVL!9}\color{domainVL}\benchmarkheader{ReasonMap} &
\cellcolor{domainVL!9}\color{domainVL}\benchmarkheader{BabyVision} &
\cellcolor{domainCost!9}\color{domainCost}{\scriptsize\textbf{MAS}} & \cellcolor{domainCost!9}\color{domainCost}{\scriptsize\textbf{Memory}} &
\cellcolor{domainCost!9}\color{domainCost}{\scriptsize\textbf{MAS}} & \cellcolor{domainCost!9}\color{domainCost}{\scriptsize\textbf{Memory}} \\
\midrule

\noalign{\gdef\mainrowshade{4}}

   &    &  No Memory
   & 15.7 & 79.5 & 57.1 & 53.0 & 66.2 & 66.4 & 20.4 & 81.0 & 49.2 & 40.1 & 14.2
   & 1.00 & 0.00 & 1.00 & 0.00 \\

   &   &  Mem0 (Qwen3.8-27B)
   & 1.4 & \underline{81.6} & 61.9 & 61.3 & 70.6 & 78.0 & 24.4 & 80.6 & 48.6 & 39.6 & 12.2
   & 1.33 & 0.31 & 1.23 & 0.11  \\

   &   &  Cognee (Qwen3.8-27B)
   & 14.3 & 73.7 & 60.1 & \underline{61.5} & 69.9 & 78.1 & 44.2 & 78.2 & 51.2 & 35.4 & 11.5
   & 1.16 & 1.66 & 1.11 & 0.75 \\

   &   &  A-Mem (Qwen3.8-27B)
   & 12.9 & 80.2 & 66.3 & 60.0 & \underline{72.3} & \underline{78.6} & 25.4 & 81.8 & 45.4 & 40.6 & 13.5
   & 1.23 & 0.24 & 1.27 & 0.81 \\

   &   & Agent-KB (Qwen3.8-27B)
   & 11.4 & 78.6 & 59.5 & 25.8 & 67.4 & 69.8 & 25.2 & 80.8 & 45.0 & 24.1 & \underline{15.6}
   & 1.54 & 0.31 & 1.12 & 0.14 \\

\noalign{\gdef\mainrowshade{12}}
   &   &  \textbf{EpiCon (ours 2B)}
   & \underline{18.6} & 66.1 & \underline{67.3} & 59.9 & 71.4 & 75.2 & \underline{45.2} & \underline{84.2} & \underline{52.4} & \underline{41.5} & 15.3
   & 1.16 & 0.30 & 1.04 & 0.67  \\

\noalign{\gdef\mainrowshade{16}}
   & \multirow{-7}{*}{Qwen3.8-27B} & \textbf{EpiCon (Qwen3.8-27B)}
   & \textbf{22.9} & \textbf{83.5} & \textbf{68.5} & \textbf{61.7} & \textbf{73.0} & \textbf{79.3}
   & \textbf{48.6} & \textbf{85.4} & \textbf{53.8} & \textbf{43.4} & \textbf{16.3}
   & 1.15 & 0.92 & 1.03 & 0.64 \\

\cmidrule(lr){2-18}

\noalign{\gdef\mainrowshade{4}}
  & 
  & No Memory
  & 5.7 & 46.2 & 43.3 & 51.7 & 65.9 & 65.6 & 25.8 & 85.0
  & 45.4 & 31.6 & 17.7
  & 1.00 & 0.00 & 1.00 & 0.00 \\

  & & Mem0 (Gemma4-31B)
  & \underline{7.1} & 54.0 & 33.4 & 51.2 & 65.1 & 62.2 & 29.2 & 86.2
  & 44.4 & 27.8 & 12.2
  & 1.40 & 0.53 & 1.24 & 0.12 \\

  & & Cognee (Gemma4-31B)
  & 4.3 & 49.9 & 27.6 & 54.6 & 49.0 & 63.7 & 37.4 & 72.4
  & 42.2 & 27.4 & 12.8
  & 0.92 & 7.23 & 1.07 & 1.02 \\

  & & A-Mem (Gemma4-31B)
  & \textbf{8.6} & 55.5 & 37.0 & \underline{54.9} & \textbf{68.6} & 63.1 & 29.8 & \underline{88.0}
  & 44.6 & 31.6 & 13.9
  & 1.23 & 0.39 & 1.31 & 0.97 \\

  &   & Agent-KB (Gemma4-31B)
  & 5.7 & 51.4 & 27.4 & 52.8 & 62.1 & 45.9 & 26.2 & 87.2 & 46.8 & \underline{32.5} & 16.3 & 1.41 & 0.31 & 1.04 & 0.13 \\

\noalign{\gdef\mainrowshade{12}}
  & & \textbf{EpiCon (ours 2B)}
  & \textbf{8.6} & \underline{58.8} & \underline{44.2} & 54.3 & 67.5 & \textbf{66.7}
  & \underline{38.8} & 87.8 & \underline{53.2} & 32.1 & \underline{18.1}
  & 1.33 & 0.47 & 1.01 & 0.74 \\

\noalign{\gdef\mainrowshade{16}}
  \multirow{-14}{*}{\textbf{Codex}} & \multirow{-7}{*}{Gemma4-31B} & \textbf{EpiCon (Gemma4-31B)}
  & \textbf{8.6} & \textbf{59.6} & \textbf{45.8} & \textbf{57.1} & \underline{67.8} & \underline{66.4}
  & \textbf{57.6} & \textbf{88.4} & \textbf{54.2} & \textbf{36.3} & \textbf{19.1}
  & 1.20 & 1.56 & 1.02 & 0.67 \\

\midrule

\noalign{\gdef\mainrowshade{4}}

   &    &  No Memory
   & 14.3 & 79.5 & 52.4 & 53.7 & 61.4 & 65.3 & 42.2 & 88.8 & 54.0 & 22.6 & 20.1
   & 1.00 & 0.00 & 1.00 & 0.00 \\

   &   &  Mem0 (Qwen3.8-27B)
   & 11.4 & 80.5 & 46.1 & 54.9 & 61.0 & \textbf{68.8} & 38.8 & 90.0 & 52.2 & 21.2 & 17.0
   & 1.10 & 0.15 & 1.87 & 0.41 \\

   &   &  Cognee (Qwen3.8-27B)
   & 15.7 & 76.6 & 50.9 & 54.0 & 60.6 & \underline{68.3} & 53.6 & 91.2 & 50.2 & 19.3 & 18.4
   & 0.95 & 0.85 & 1.33 & 2.70 \\

   &   &  A-Mem (Qwen3.8-27B)
   & 17.1 & 80.6 & \underline{55.4} & 53.7 & \underline{63.4} & 65.5 & 43.6 & 89.6 & 52.2 & 22.2 & 17.4
   & 1.09 & 0.12 & 2.01 & 2.98 \\

   &   & Agent-KB (Qwen3.8-27B)
   & \textbf{20.0} & 79.5 & 50.5 & 52.5 & 62.6 & 63.8 & 42.6 & 90.4 & 52.6 & 13.7 & \underline{20.5} & 1.13 & 0.15 & 1.51 & 0.64 \\

\noalign{\gdef\mainrowshade{12}}
   &   &  \textbf{EpiCon (ours 2B)}
   & \underline{18.6} & \underline{81.8} & 53.6 & \underline{56.9} & \textbf{63.5} & 65.2 & \underline{56.6} & \textbf{92.2} & \underline{57.4} & \underline{24.5} & 17.4
   & 1.08 & 0.15 & 1.06 & 1.92 \\

\noalign{\gdef\mainrowshade{16}}
   & \multirow{-7}{*}{Qwen3.8-27B} & \textbf{EpiCon (Qwen3.8-27B)}
   & \textbf{20.0} & \textbf{82.9} & \textbf{57.0} & \textbf{61.2} & 62.8 & 65.4 & \textbf{58.2} & \underline{91.8} & \textbf{58.2} & \textbf{27.4} & \textbf{23.3}
   & 1.07 & 0.57 & 1.10 & 2.88 \\

\cmidrule(lr){2-18}

\noalign{\gdef\mainrowshade{4}}
  & 
  & No Memory
  & \underline{10.0} & 2.7 & 35.6 & 51.4 & 55.4 & 32.8 & 16.0 & 91.0
  & 49.6 & 27.4 & 11.5
  & 1.00 & 0.00 & 1.00 & 0.00 \\

  & & Mem0 (Gemma4-31B)
  & 7.1 & \underline{9.1} & 35.7 & 51.6 & 51.6 & 32.6 & 11.4 & 88.6
  & 49.2 & 26.4 & 12.5
  & 0.93 & 0.20 & 1.80 & 0.38 \\

  & & Cognee (Gemma4-31B)
  & 5.7 & 0.5 & 37.7 & 49.7 & 44.8 & 29.9 & 12.6 & 89.4
  & 49.8 & \textbf{30.7} & 12.8
  & 0.82 & 2.23 & 1.22 & 3.06 \\

  & & A-Mem (Gemma4-31B)
  & 7.1 & \textbf{10.8} & 38.8 & 51.0 & 52.4 & \underline{33.0} & \textbf{20.2} & 89.6
  & 52.2 & 24.1 & 12.8
  & 0.87 & 0.12 & 1.99 & 3.02 \\

   &   & Agent-KB (Gemma4-31B)
   & 7.1 & 1.8 & 36.1 & 50.0 & 51.5 & 29.5 & 13.0 & 89.2 & 47.8 & 25.5 & 15.6 & 0.86 & 0.18 & 1.38 & 0.60 \\

\noalign{\gdef\mainrowshade{12}}
  & & \textbf{EpiCon (ours 2B)}
  & 5.7 & 2.0 & \underline{41.0} & \underline{52.9} & \textbf{57.5} & \textbf{34.3}
  & \underline{18.8} & \underline{91.2} & \textbf{54.0} & 28.8 & \underline{16.3}
  & 0.86 & 0.14 & 1.08 & 1.70 \\

\noalign{\gdef\mainrowshade{16}}
  \multirow{-14}{*}{\shortstack[l]{\textbf{Deepseek}\\\textbf{Harness}}} & \multirow{-7}{*}{Gemma4-31B} & \textbf{EpiCon (Gemma4-31B)}
  & \textbf{11.4} & 0.7 & \textbf{46.3} & \textbf{54.3} & \underline{56.8} & 32.9
  & 18.4 & \textbf{92.2} & \underline{53.2} & \underline{29.2} & \textbf{17.4}
  & 0.86 & 0.53 & 1.09 & 2.25 \\

\bottomrule

\end{tabular}%
}
\gdef\mainrowshade{4}
\endgroup
\vspace{-6mm}
\end{table*}

\subsection{Question-Level Multimodal Memory Evolution}
\label{sec:method:controller}

\textbf{Textual and visual memory co-evolution.}
Question-level memory is represented as $W_{i,t}=(M^{\mathrm{text}}_{i,t},M^{\mathrm{vis}}_{i,t})$. Text records actionable guidance, applicability conditions, and cautions; visual memory points to supporting regions in source images. When another attempt is available, the controller updates memory using the previous state, current question and images, latest answer, and execution trace:
\begin{equation}
W_{i,t}=C_\theta^{\mathrm{update}}(W_{i,t-1},x_i,\hat y_{i,t},\tau_{i,t}).
\label{eq:epicon-update}
\end{equation}
The equation denotes the accepted update after harness validation. The controller jointly proposes revised text, a source image, and a region. It can retain, replace, or remove visual evidence as the guidance evolves. The harness validates the proposal and generates the requested crop; invalid proposals leave memory unchanged. The accepted text and visual evidence guide the next attempt.

\textbf{Adaptive visual memory injection.}
The controller decides whether visual memory should be included in the next input through $g_{i,t}\in\{0,1\}$. For retrieved experience, the decision is based on the current question and descriptions of the retrieved memory. For question-level memory, the joint update supplies the decision together with the revised text and region. Let $T_{i,t}$ and $V_{i,t}$ denote the text and available visual memory for attempt $t$, drawn from retrieved experience or the latest question-level state. The memory input is
\begin{equation}
c_{i,t}=\begin{cases}
(T_{i,t},\varnothing), & g_{i,t}=0,\\
(T_{i,t},V_{i,t}), & g_{i,t}=1.
\end{cases}
\label{eq:epicon-visual-injection}
\end{equation}
Here $V_{i,t}$ is bounded by the memory-image budget. The gate selects input content without deleting stored visual memory. The current question's original images remain available in both cases.

\begin{figure*}[t]
\vspace{-2mm}
\centering
\includegraphics[width=\textwidth]{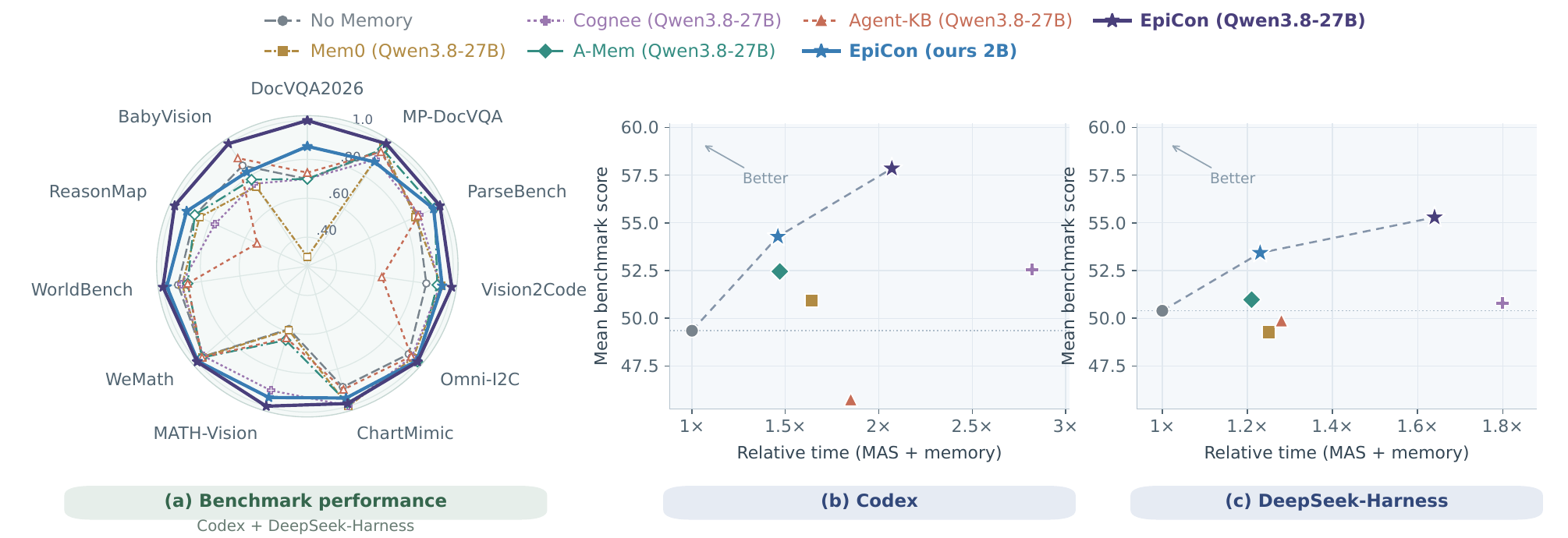}
\vspace{-3mm}
\caption{Benchmark profiles and performance--time trade-offs with Qwen3.8-27B. (a) Scores are averaged across the two harnesses and normalized by the best method mean per benchmark; the radial axis starts at 0.25. (b) and (c) show scores averaged equally over eleven benchmarks; time sums normalized MAS and memory costs. Dashed lines connect No Memory and the two EpiCon variants as visual guides.}
\label{fig:benchmark_cost_maxratio_zoom}
\vspace{-6mm}
\end{figure*}

\subsection{Shared Experience Consolidation and Reuse}
\label{sec:method:tree}

\textbf{Organizing accumulated lessons.}
The shared bank uses a tree to organize lessons with their guidance, source references, and visual evidence. Leaves store question-specific lessons, while internal nodes group and summarize related experience. During maintenance, the Tree Self-Organizer can propose Place, Merge, Split + Lift, and Consolidation operations to assign categories, combine related groups, separate broad groups, and summarize shared procedures and conditions. The harness validates and executes these proposals and can enforce capacity-based partitions.

\textbf{Abstracting reusable rules.}
Consolidation captures common guidance and conflicts across lessons. The harness determines abstraction levels and rule eligibility from support, failures, conflicts, and evidence strength, while the organizer generates summaries. Internal nodes can retain qualified summaries or become reusable rules when the evidence permits. This organization allows later tasks to access specific experience and guidance consolidated from multiple lessons.

\textbf{Retrieving experience for subsequent tasks.}
\label{sec:method:retrieval}
Retrieval first recalls candidates and then selects relevant experience:
\begin{equation}
R_i=S_\phi^{\mathrm{retrieve}}\!\left(x_i,\operatorname{Recall}(x_i,B)\right).
\label{eq:epicon-retrieval}
\end{equation}
Here $\operatorname{Recall}$ uses a frozen text encoder to retrieve candidates from node indices describing their topic, scope, and question anchor. The organizer examines the current question and candidate memories, including textual guidance and associated visual evidence. It selects applicable rules, a concrete experience, or nothing when no candidate applies. The selected content is available to the receiving MAS regardless of which harness produced it, with visual delivery controlled by $C_\theta$. Retrieval uses existing bank content; further construction or expansion can add lessons and revise its organization for subsequent use. Figure~\ref{fig:metro_memory_showcase} illustrates how question-level memory co-evolution connects to experience consolidation and reuse across transit maps.
\vspace{-2mm}

\section{Experiments}
\label{sec:experiments}

\subsection{Experimental Setup}
\label{sec:exp:setup}

\textbf{Baselines and model configurations.}
Our main experiments use two multi-agent system (MAS) harnesses, Codex~\citep{openai2026codex} and DeepSeek-Harness~\citep{deepseek-harness2026}, each paired with Qwen3.8-27B~\citep{qwen38} and Gemma4-31B~\citep{gemmateam2026gemma4}. We additionally evaluate Codex with GPT-5.6-Luna~\citep{openai2026gpt56luna}. We compare EpiCon with No Memory, Mem0~\citep{chhikara2025mem0}, Cognee~\citep{markovic2025cognee}, A-Mem~\citep{xu2026mem}, and Agent-KB~\citep{tang2025agent}. External memory systems use the corresponding MAS backbone zero-shot for memory operations. Backbone-sized EpiCon uses the same backbone for memory control and tree organization; the 2B variant uses two independently trained 2B models.

\textbf{Evaluation suite.}
We evaluate 4,538 questions from 11 benchmarks spanning document understanding, visual-to-code generation, vision-grounded mathematics, and general visual-language reasoning: DocVQA2026~\citep{docvqa2026}, MP-DocVQA~\citep{tito2022hierarchical}, ParseBench~\citep{zhang2026parsebench}, Vision2Code~\citep{periasami2026vision2code}, Omni-I2C~\citep{zhou2026omni}, ChartMimic~\citep{yang2025chartmimic}, MATH-Vision~\citep{wang2024measuring}, WeMath~\citep{qiao2025we}, WorldBench~\citep{yin2026worldbench}, ReasonMap~\citep{feng2026reasonmap}, and BabyVision~\citep{chen2026babyvision}.

\textbf{Evaluation protocol.}
Across all experiments, No Memory makes a single solving attempt per question. For the main comparison, each memory-based method builds its own bank using a shared set of 1,000 construction questions from the same benchmarks, disjoint from the evaluation questions under our project-defined splits. ReasonMap is split by map, with no map shared between construction and evaluation. All evaluations with historical memory retrieval use a single solving attempt per question, with question-level memory updates disabled. In Table~\ref{tab:main_results}, this single-attempt protocol applies to every method. Each attempt follows the harness's standard internal workflow. Model parameters and historical banks remain frozen during evaluation. Reference answers are used only for offline scoring. We report scores, time, and token usage, with MAS and memory costs separated. Time and token usage are normalized to the corresponding No Memory MAS costs. All experiments are run using NVIDIA H100 GPUs; GPT-5.6-Luna is accessed through its API.

\subsection{Main Results}
\label{sec:exp:main}
\begin{table*}[t]
\vspace{-4mm}
\centering
\caption{Cross-backbone and cross-harness experience transfer from a bank built by Codex with Qwen3.8-27B. The evaluated MAS performs one solve; EpiCon (ours 2B) reuses the frozen bank without question-level updates.}
\label{tab:experience_transfer}
\begingroup
\setlength{\aboverulesep}{0pt}
\setlength{\belowrulesep}{0pt}
\setlength{\cmidrulesep}{0pt}
\setlength{\minrowclearance}{1.5pt}
\providecommand{\mainrowshade}{4}
\definecolor{domainDoc}{HTML}{35664D}
\definecolor{domainCode}{HTML}{005A53}
\definecolor{domainMath}{HTML}{00536B}
\definecolor{domainVL}{HTML}{32487A}
\newcolumntype{D}{>{\cellcolor{domainDoc!\mainrowshade}}c}
\newcolumntype{V}{>{\cellcolor{domainCode!\mainrowshade}}c}
\newcolumntype{M}{>{\cellcolor{domainMath!\mainrowshade}}c}
\newcolumntype{R}{>{\cellcolor{domainVL!\mainrowshade}}c}

\setlength{\tabcolsep}{3.4pt}
\renewcommand{\arraystretch}{1.10}
\newcommand{\benchmarkheader}[1]{%
  \rotatebox[origin=lb]{60}{\strut\textbf{\scriptsize #1}}}
\resizebox{\textwidth}{!}{%
\begin{tabular}{lll DDD VVV MM RRR}
\toprule
\shortstack[c]{\textbf{MAS}\\\textbf{Harness}} &
\shortstack[c]{\textbf{MAS}\\\textbf{Backbone}} &
\textbf{Memory System} &
\multicolumn{3}{c}{\cellcolor{domainDoc!9}\color{domainDoc}\shortstack[c]{\textbf{Document}\\\textbf{Understanding}}} &
\multicolumn{3}{c}{\cellcolor{domainCode!9}\color{domainCode}\textbf{Visual-to-Code}} &
\multicolumn{2}{c}{\cellcolor{domainMath!9}\color{domainMath}\shortstack[c]{\textbf{Vision-Grounded}\\\textbf{Math}}} &
\multicolumn{3}{c}{\cellcolor{domainVL!9}\color{domainVL}\shortstack[c]{\textbf{General VL}\\\textbf{Reasoning}}} \\
& & &
\cellcolor{domainDoc!9}\color{domainDoc}\benchmarkheader{DocVQA2026} &
\cellcolor{domainDoc!9}\color{domainDoc}\benchmarkheader{MP-DocVQA} &
\cellcolor{domainDoc!9}\color{domainDoc}\benchmarkheader{ParseBench} &
\cellcolor{domainCode!9}\color{domainCode}\benchmarkheader{Vision2Code} &
\cellcolor{domainCode!9}\color{domainCode}\benchmarkheader{Omni-I2C} &
\cellcolor{domainCode!9}\color{domainCode}\benchmarkheader{ChartMimic} &
\cellcolor{domainMath!9}\color{domainMath}\benchmarkheader{MATH-Vision} &
\cellcolor{domainMath!9}\color{domainMath}\benchmarkheader{WeMath} &
\cellcolor{domainVL!9}\color{domainVL}\benchmarkheader{WorldBench} &
\cellcolor{domainVL!9}\color{domainVL}\benchmarkheader{ReasonMap} &
\cellcolor{domainVL!9}\color{domainVL}\benchmarkheader{BabyVision} \\
\midrule
\noalign{\gdef\mainrowshade{4}}
\multirow{2}{*}{Codex} & \multirow{2}{*}{Gemma4-31B} & No Memory
& \textbf{5.7} & 46.2 & 43.3 & 51.7 & 65.9 & 65.6 & 25.8 & \textbf{85.0}
& 45.4 & 31.6 & 17.7\\
\noalign{\gdef\mainrowshade{12}}
& & \textbf{EpiCon (ours 2B)}
& \textbf{5.7} & \textbf{60.4} & \textbf{45.3} & \textbf{53.8} & \textbf{67.8} & \textbf{68.2} & \textbf{33.0} & \textbf{85.0} & \textbf{47.8} & \textbf{34.4} & \textbf{18.1} \\
\midrule
\noalign{\gdef\mainrowshade{4}}
\multirow{2}{*}{\shortstack[l]{DeepSeek\\Harness}} & \multirow{2}{*}{Qwen3.8-27B} & No Memory
& 14.3 & \textbf{79.5} & 52.4 & 53.7 & 61.4 & 65.3 & 42.2 & 88.8 & \textbf{54.0} & 22.6 & 20.1 \\
\noalign{\gdef\mainrowshade{12}}
& & \textbf{EpiCon (ours 2B)}
& \textbf{18.6} & 79.1 & \textbf{65.7} & \textbf{59.2} & \textbf{63.3} & \textbf{68.7} & \textbf{56.0} & \textbf{90.6} & 52.2 & \textbf{26.4} & \textbf{20.5} \\
\bottomrule
\end{tabular}%
}
\gdef\mainrowshade{4}
\endgroup
\vspace{-4mm}
\end{table*}

\begin{table*}[t]
\vspace{-2mm}
\centering
\caption{Cross-harness memory evolution with EpiCon (ours 2B). Each evaluated harness builds the original bank; the other harness uses and evolves it. The two stages each use 1,000 disjoint construction questions. Both bank conditions use a single solving attempt.}
\label{tab:experience_transfer_roundtrip}
\begingroup
\setlength{\aboverulesep}{0pt}
\setlength{\belowrulesep}{0pt}
\setlength{\cmidrulesep}{0pt}
\setlength{\minrowclearance}{1.5pt}
\providecommand{\mainrowshade}{4}
\definecolor{domainDoc}{HTML}{35664D}
\definecolor{domainCode}{HTML}{005A53}
\definecolor{domainMath}{HTML}{00536B}
\definecolor{domainVL}{HTML}{32487A}
\newcolumntype{D}{>{\cellcolor{domainDoc!\mainrowshade}}c}
\newcolumntype{V}{>{\cellcolor{domainCode!\mainrowshade}}c}
\newcolumntype{M}{>{\cellcolor{domainMath!\mainrowshade}}c}
\newcolumntype{R}{>{\cellcolor{domainVL!\mainrowshade}}c}

\setlength{\tabcolsep}{3.4pt}
\renewcommand{\arraystretch}{1.10}
\newcommand{\benchmarkheader}[1]{%
  \rotatebox[origin=lb]{60}{\strut\textbf{\scriptsize #1}}}
\resizebox{\textwidth}{!}{%
\begin{tabular}{lll DDD VVV MM RRR}
\toprule
\shortstack[c]{\textbf{MAS}\\\textbf{Harness}} &
\shortstack[c]{\textbf{MAS}\\\textbf{Backbone}} &
\textbf{Memory Bank} &
\multicolumn{3}{c}{\cellcolor{domainDoc!9}\color{domainDoc}\shortstack[c]{\textbf{Document}\\\textbf{Understanding}}} &
\multicolumn{3}{c}{\cellcolor{domainCode!9}\color{domainCode}\textbf{Visual-to-Code}} &
\multicolumn{2}{c}{\cellcolor{domainMath!9}\color{domainMath}\shortstack[c]{\textbf{Vision-Grounded}\\\textbf{Math}}} &
\multicolumn{3}{c}{\cellcolor{domainVL!9}\color{domainVL}\shortstack[c]{\textbf{General VL}\\\textbf{Reasoning}}} \\
& & &
\cellcolor{domainDoc!9}\color{domainDoc}\benchmarkheader{DocVQA2026} &
\cellcolor{domainDoc!9}\color{domainDoc}\benchmarkheader{MP-DocVQA} &
\cellcolor{domainDoc!9}\color{domainDoc}\benchmarkheader{ParseBench} &
\cellcolor{domainCode!9}\color{domainCode}\benchmarkheader{Vision2Code} &
\cellcolor{domainCode!9}\color{domainCode}\benchmarkheader{Omni-I2C} &
\cellcolor{domainCode!9}\color{domainCode}\benchmarkheader{ChartMimic} &
\cellcolor{domainMath!9}\color{domainMath}\benchmarkheader{MATH-Vision} &
\cellcolor{domainMath!9}\color{domainMath}\benchmarkheader{WeMath} &
\cellcolor{domainVL!9}\color{domainVL}\benchmarkheader{WorldBench} &
\cellcolor{domainVL!9}\color{domainVL}\benchmarkheader{ReasonMap} &
\cellcolor{domainVL!9}\color{domainVL}\benchmarkheader{BabyVision} \\
\midrule
\noalign{\gdef\mainrowshade{4}}
\multirow{2}{*}{Codex} & \multirow{2}{*}{Qwen3.8-27B} & Original bank
& \textbf{18.6} & 66.1 & \textbf{63.4} & 59.9 & \textbf{71.4} & 75.2 & 45.2 & 84.2 & 52.4 & 38.3 & 14.3 \\
\noalign{\gdef\mainrowshade{12}}
& & \textbf{Evolved bank}
& \textbf{18.6} & \textbf{80.3} & 58.9 & \textbf{61.3} & 70.9 & \textbf{76.7} & \textbf{45.4} & \textbf{88.4} & \textbf{55.2} & \textbf{40.1} & \textbf{16.8} \\

\midrule
\noalign{\gdef\mainrowshade{4}}
\multirow{2}{*}{\shortstack[l]{DeepSeek\\Harness}} & \multirow{2}{*}{Qwen3.8-27B} & Original bank
& 18.6 & 81.8 & 52.3 & 56.9 & 63.5 & 65.2 & \textbf{56.6} & \textbf{92.2} & 57.4 & 23.5 & 19.3 \\
\noalign{\gdef\mainrowshade{12}}
& & \textbf{Evolved bank}
& \textbf{20.0} & \textbf{82.7} & \textbf{65.9} & \textbf{58.7} & \textbf{69.2} & \textbf{72.1} & 53.4 & 91.2 & \textbf{57.6} & \textbf{29.6} & \textbf{22.3} \\
\bottomrule
\end{tabular}%
}
\gdef\mainrowshade{4}
\endgroup
\vspace{-6mm}
\end{table*}

With a single solving attempt per question, backbone-sized EpiCon achieves the highest eleven-benchmark macro-average in all four host configurations, exceeding the strongest external memory baseline in each by 1.9 to 5.9 points (Table~\ref{tab:main_results}). Macro-averages give equal weight to each benchmark. For Codex with Qwen3.8-27B, this variant improves ParseBench from 57.1 to 68.5 and MATH-Vision from 20.4 to 48.6 over No Memory. Averaged over the two Qwen3.8-27B host configurations, it also scores highest on ten of eleven benchmarks (Figure~\ref{fig:benchmark_cost_maxratio_zoom}(a)).

Replacing backbone-sized memory models with our two trained 2B models reduces memory-operation time by approximately 67\% to 74\% and total (MAS + memory) time by 25\% to 35\%, with macro-average scores lower by 0.9 to 3.6 points. The 2B variant still improves macro-average scores over No Memory by 1.7 to 4.9 points across the four host configurations. In Figure~\ref{fig:benchmark_cost_maxratio_zoom}(b) and (c), both EpiCon configurations lie on the empirical Pareto front among the compared methods for each harness with Qwen3.8-27B.

Our two 2B memory models are trained on supervision from three domains: visual mathematics, document understanding, and reasoning over charts and infographics (Section~\ref{sec:data}). They also support gains on general visual-language reasoning: the 2B variant improves WorldBench in all four configurations ($+3.2$ to $+7.8$ points), while changes on ReasonMap ($+0.5$ to $+1.9$) and BabyVision ($-2.7$ to $+4.8$) are smaller and less consistent. These results suggest that the trained memory models can manage experience from additional task types without further parameter updates.

\subsection{Experience Transfer and Memory Expansion}
\label{sec:exp:experience_transfer}

We evaluate experience reuse across harnesses and backbones, followed by continued accumulation in a shared bank. All comparisons use EpiCon with fixed Memory Controller (MC) and Tree Self-Organizer (Tree) models.

\textbf{Cross-backbone and cross-harness transfer.}
A bank built by Codex with Qwen3.8-27B is reused by Codex with Gemma4-31B and DeepSeek-Harness with Qwen3.8-27B, changing only the backbone or the harness, respectively. Each evaluated MAS performs one solve with or without the frozen bank, disabling question-level updates to assess historical experience reuse. Table~\ref{tab:experience_transfer} compares performance across all eleven benchmarks, with macro-average improvements of 3.2 and 4.2 points for backbone and harness transfer, respectively. For example, backbone transfer improves MATH-Vision from 25.8 to 33.0, while harness transfer improves ParseBench from 52.4 to 65.7. These gains show that a system can benefit from existing shared experience without first constructing its own bank, even when its harness or backbone differs from that of the original contributor.

\textbf{Cross-harness memory evolution.}
We test whether experience contributed by another harness can benefit the harness that built the bank. One harness builds the original bank on 1,000 construction questions; the other uses and evolves it on a separate set of 1,000 questions. We evaluate both directions between Codex and DeepSeek-Harness, with Qwen3.8-27B as the backbone throughout. In each direction, the harness that built the original bank is evaluated with the original and evolved banks on the same 4,388 questions across eleven benchmarks, disjoint from both construction sets. Both conditions use one retrieval and a single solving attempt, without question-level memory updates. Table~\ref{tab:experience_transfer_roundtrip} compares the original and evolved banks. Macro-average scores rise from 53.5 to 55.7 for Codex and from 53.4 to 56.6 for DeepSeek-Harness, gains of 2.1 and 3.2 points. The evolved banks improve eight and nine of eleven benchmark scores, respectively.  The comparison measures the overall effect of bank evolution and expansion; construction splits are provided in Appendix~\ref{sec:appendix:roundtrip}.

The benefits of cross-harness evolution depend on the task and contributing system. On ParseBench, where Codex is stronger in Table~\ref{tab:main_results}, its score drops from 63.4 to 58.9 after DeepSeek-Harness evolves its bank, while DeepSeek-Harness improves from 52.3 to 65.9 after Codex evolves its bank. On MATH-Vision, Codex changes only slightly from 45.2 to 45.4, while DeepSeek-Harness declines from 56.6 to 53.4. These task-dependent patterns suggest that experience from different systems can provide complementary guidance, although individual benchmarks do not improve uniformly. Taken together, Tables~\ref{tab:experience_transfer} and~\ref{tab:experience_transfer_roundtrip} support collective learning through shared memory: a system can benefit from experience accumulated by another system, then contribute additional experience that improves subsequent solving by the original contributor.
\begin{table*}[t]
\vspace{-4mm}
\centering
\caption{Question-level memory control with no memory bank. No Memory makes a single solving attempt; MC settings allow up to five attempts. \textbf{MC (Qwen3.8-27B)} uses Qwen3.8-27B as a zero-shot memory controller and \textbf{MC (ours 2B)} uses our trained 2B controller. We report task performance on four benchmarks. Time and token usage are normalized separately within each MAS harness using the corresponding No Memory MAS cost as the $1\times$ baseline; the same baseline normalizes memory costs.}
\label{tab:question_level_mc}
\begingroup
\setlength{\aboverulesep}{0pt}
\setlength{\belowrulesep}{0pt}
\setlength{\cmidrulesep}{0pt}
\setlength{\minrowclearance}{1.5pt}
\providecommand{\mainrowshade}{4}
\definecolor{tablePerf}{HTML}{00536B}
\definecolor{tableCost}{HTML}{493F7B}
\newcolumntype{P}{>{\cellcolor{tablePerf!\mainrowshade}}c}
\newcolumntype{C}{>{\cellcolor{tableCost!\mainrowshade}}c}

\footnotesize
\setlength{\tabcolsep}{4pt}
\renewcommand{\arraystretch}{1.15}
\resizebox{\textwidth}{!}{%
\begin{tabular}{lll PPPP CCCC}
\toprule
\textbf{MAS} &
\textbf{MAS Backbone} &
\textbf{Memory Setting} &
\multicolumn{4}{c}{\cellcolor{tablePerf!9}\color{tablePerf}\textbf{Task Performance}} &
\multicolumn{2}{c}{\cellcolor{tableCost!9}\color{tableCost}\textbf{Time ($\times$)}} &
\multicolumn{2}{c}{\cellcolor{tableCost!9}\color{tableCost}\textbf{Tokens ($\times$)}} \\

& & &
\cellcolor{tablePerf!9}\color{tablePerf}\shortstack{\scriptsize Document\\\scriptsize ParseBench} &
\cellcolor{tablePerf!9}\color{tablePerf}\shortstack{\scriptsize Visual-to-Code\\\scriptsize Vision2Code} &
\cellcolor{tablePerf!9}\color{tablePerf}\shortstack{\scriptsize Vision Math\\\scriptsize MATH-Vision} &
\cellcolor{tablePerf!9}\color{tablePerf}\shortstack{\scriptsize General VL\\\scriptsize BabyVision} &
\cellcolor{tableCost!9}\color{tableCost}\textbf{MAS} & \cellcolor{tableCost!9}\color{tableCost}\textbf{Memory} & \cellcolor{tableCost!9}\color{tableCost}\textbf{MAS} & \cellcolor{tableCost!9}\color{tableCost}\textbf{Memory} \\
\midrule

\noalign{\gdef\mainrowshade{4}}
\multirow{3}{*}{Codex} & \multirow{3}{*}{Qwen3.8-27B} & No Memory
& 57.1 & 53.0 & 20.4 & \underline{14.2}
& 1.00 & 0.00 & 1.00 & 0.00 \\

\noalign{\gdef\mainrowshade{12}}
& & \textbf{MC (ours 2B)}
& \underline{59.8} & \underline{55.9} & \underline{41.2} & \textbf{17.7}
& 2.16 & 0.23 & 2.40 & 0.96 \\

\noalign{\gdef\mainrowshade{16}}
& & \textbf{MC (Qwen3.8-27B)}
& \textbf{63.1} & \textbf{56.6} & \textbf{49.6} & 13.9
& 2.57 & 0.23 & 3.07 & 0.95 \\

\midrule

\noalign{\gdef\mainrowshade{4}}
\multirow{3}{*}{\shortstack[l]{DeepSeek\\Harness}}
& \multirow{3}{*}{Qwen3.8-27B} & No Memory
& 52.4 & 53.7 & 42.2 & 20.1
& 1.00 & 0.00 & 1.00 & 0.00 \\

\noalign{\gdef\mainrowshade{12}}
& & \textbf{MC (ours 2B)}
& \underline{55.1} & \underline{57.8} & \underline{51.4} & \underline{23.3}
& 2.10 & 0.13 & 2.44 & 3.23 \\

\noalign{\gdef\mainrowshade{16}}
& & \textbf{MC (Qwen3.8-27B)}
& \textbf{64.5} & \textbf{65.8} & \textbf{61.6} & \textbf{35.1}
& 2.47 & 0.16 & 2.65 & 3.25 \\ 
\bottomrule
\end{tabular}%
}
\gdef\mainrowshade{4}
\endgroup
\vspace{-4mm}
\end{table*}

\subsection{Memory Ablations}
\label{sec:exp:memory_analysis}

\textbf{Question-level memory control.}
\label{sec:exp:question_level}
With historical retrieval disabled, Table~\ref{tab:question_level_mc} compares No Memory, MC (ours 2B), and MC (Qwen3.8-27B). No Memory makes a single solving attempt, while MC revises memory across attempts using the question, images, existing memory, answer, and execution trace. Refinement stops after at most five attempts or earlier when the system judges the answer ready to return. With the trained 2B controller, refinement improves all four scores under both harnesses, including Vision2Code from 53.7 to 57.8 with DeepSeek-Harness. The larger controller scores higher in seven of eight comparisons, while the 2B controller requires less MAS execution time and no more memory-operation time.

\textbf{Visual memory and adaptive injection.}
\label{sec:exp:multimodal}
Table~\ref{tab:multimodal_ablation} fixes MC (ours 2B), disables historical retrieval, and uses the same budget of up to five attempts across memory-enabled settings. No Memory makes a single solving attempt; original task images remain available throughout. With adaptive injection fixed, evolving visual memory improves seven scores over frozen memory and ties the remaining score, supporting the benefit of memory co-evolution beyond selective injection. With always-on injection, frozen visual memory improves Codex's MATH-Vision score from 27.2 to 40.6 over text-only memory. Adaptive injection improves seven of eight scores with either frozen or evolving visual memory. For evolving memory, it also reduces MAS execution time by approximately 25\% with Codex and 17\% with DeepSeek-Harness, with lower MAS token usage under both harnesses.

\begin{table*}[t]
\vspace{-4mm}
\centering
\caption{
Multimodal memory ablation with trained MC 2B and no historical bank.
Memory-enabled settings share a budget of up to five attempts; the row with three dashes in the memory configuration columns denotes No Memory with a single solving attempt. Visual memory can be frozen or evolving, and is
injected either on every attempt (Always) or adaptively (Adaptive).
Original task images remain available in every setting.
}
\label{tab:multimodal_ablation}

\begingroup
\setlength{\aboverulesep}{0pt}
\setlength{\belowrulesep}{0pt}
\setlength{\cmidrulesep}{0pt}
\setlength{\minrowclearance}{1.5pt}
\providecommand{\mainrowshade}{4}
\definecolor{tableConfig}{HTML}{35664D}
\definecolor{tablePerf}{HTML}{00536B}
\definecolor{tableCost}{HTML}{493F7B}
\newcolumntype{G}{>{\cellcolor{tableConfig!\mainrowshade}}c}
\newcolumntype{P}{>{\cellcolor{tablePerf!\mainrowshade}}c}
\newcolumntype{C}{>{\cellcolor{tableCost!\mainrowshade}}c}

\footnotesize
\setlength{\tabcolsep}{3.8pt}
\renewcommand{\arraystretch}{1.15}

\newcommand{\cmark}{\textbf{\checkmark}}
\newcommand{\xmark}{--}

\resizebox{\textwidth}{!}{%
\begin{tabular}{ll GGG PPPP CCCC}
\toprule
\textbf{MAS} &
\textbf{MAS Backbone} &
\multicolumn{3}{c}{\cellcolor{tableConfig!9}\color{tableConfig}\textbf{Memory Configuration}} &
\multicolumn{4}{c}{\cellcolor{tablePerf!9}\color{tablePerf}\textbf{Task Performance}} &
\multicolumn{2}{c}{\cellcolor{tableCost!9}\color{tableCost}\textbf{Time ($\times$)}} &
\multicolumn{2}{c}{\cellcolor{tableCost!9}\color{tableCost}\textbf{Tokens ($\times$)}} \\

& &
\cellcolor{tableConfig!9}\color{tableConfig}\shortstack{\textbf{Text}\\\textbf{Memory}} &
\cellcolor{tableConfig!9}\color{tableConfig}\shortstack{\textbf{Visual}\\\textbf{Memory}} &
\cellcolor{tableConfig!9}\color{tableConfig}\shortstack{\textbf{Visual}\\\textbf{Injection}} &
\cellcolor{tablePerf!9}\color{tablePerf}\shortstack{\scriptsize Document\\\scriptsize ParseBench} &
\cellcolor{tablePerf!9}\color{tablePerf}\shortstack{\scriptsize Visual-to-Code\\\scriptsize Vision2Code} &
\cellcolor{tablePerf!9}\color{tablePerf}\shortstack{\scriptsize Vision Math\\\scriptsize MATH-Vision} &
\cellcolor{tablePerf!9}\color{tablePerf}\shortstack{\scriptsize General VL\\\scriptsize BabyVision} &
\cellcolor{tableCost!9}\color{tableCost}\textbf{MAS} &
\cellcolor{tableCost!9}\color{tableCost}\textbf{Memory} &
\cellcolor{tableCost!9}\color{tableCost}\textbf{MAS} &
\cellcolor{tableCost!9}\color{tableCost}\textbf{Memory} \\

\midrule

\noalign{\gdef\mainrowshade{4}}
\multirow{6}{*}{Codex}
& \multirow{6}{*}{Qwen3.8-27B}
& \xmark & \xmark & \xmark
& 57.1 & 53.0 & 20.4 & 14.2
& 1.00 & 0.00 & 1.00 & 0.00 \\

\noalign{\gdef\mainrowshade{7}}
&
& \cmark & \xmark & \xmark
& \underline{58.7} & \underline{55.7} & 27.2 & 16.0
& 1.43 & 0.28 & 1.23 & 0.61 \\

\noalign{\gdef\mainrowshade{10}}
&
& \cmark & Frozen & Always
& 55.2 & 55.0 & 40.6 & 16.0
& 2.82 & 0.43 & 3.00 & 0.94 \\

\noalign{\gdef\mainrowshade{12}}
&
& \cmark & Frozen & Adaptive
& 57.1 & \textbf{55.9} & \underline{41.0} & \underline{16.3} & 2.02 & 0.25 & 2.42 & 0.94 \\

\noalign{\gdef\mainrowshade{14}}
&
& \cmark & Evolving & Always
& 55.3 & 55.2 & \underline{41.0} & \underline{16.3}
& 2.89 & 0.44 & 3.02 & 0.93 \\

\noalign{\gdef\mainrowshade{16}}
&
& \cmark & Evolving & Adaptive
& \textbf{59.8} & \textbf{55.9} & \textbf{41.2} & \textbf{17.7}
& 2.16 & 0.23 & 2.40 & 0.96 \\

\midrule

\noalign{\gdef\mainrowshade{4}}
\multirow{6}{*}{\shortstack[l]{DeepSeek\\Harness}}
& \multirow{6}{*}{Qwen3.8-27B}
& \xmark & \xmark & \xmark
& 52.4 & 53.7 & 42.2 & 20.1
& 1.00 & 0.00 & 1.00 & 0.00 \\

\noalign{\gdef\mainrowshade{7}}
&
& \cmark & \xmark & \xmark
& 51.8 & 55.1 & 47.4 & \underline{21.2}
& 1.33 & 0.13 & 1.25 & 2.37 \\

\noalign{\gdef\mainrowshade{10}}
&
& \cmark & Frozen & Always
& 51.2 & \underline{57.5} & 50.2 & 20.5
& 2.52 & 0.21 & 3.13 & 3.75 \\

\noalign{\gdef\mainrowshade{12}}
&
& \cmark & Frozen & Adaptive
& \underline{52.8} & 57.3 & 50.8 & \underline{21.2} & 1.95 & 0.13 & 2.41 & 3.58 \\

\noalign{\gdef\mainrowshade{14}}
&
& \cmark & Evolving & Always
& 52.1 & 57.0 & \textbf{51.8} & \underline{21.2}
& 2.53 & 0.21 & 3.15 & 3.72 \\

\noalign{\gdef\mainrowshade{16}}
&
& \cmark & Evolving & Adaptive
& \textbf{55.1} & \textbf{57.8} & \underline{51.4} & \textbf{23.3}
& 2.10 & 0.13 & 2.44 & 3.23 \\

\bottomrule
\end{tabular}%
}
\gdef\mainrowshade{4}
\endgroup
\vspace{-4mm}
\end{table*}

\begin{table*}[t]
\vspace{-2mm}
\centering
\begin{minipage}[t]{0.47\textwidth}
\vspace{0pt}
\caption{Memory organization with EpiCon (ours 2B). Flat and tree memory use the same source questions to create lessons.}
\label{tab:tree_memory_ablation}
\end{minipage}\hfill
\begin{minipage}[t]{0.50\textwidth}
\vspace{0pt}
\caption{Memory effectiveness on Codex with GPT-5.6-Luna using EpiCon (ours 2B). Both settings make a single solving attempt.}
\label{tab:stronger_backbones}
\end{minipage}
\par\nointerlineskip
\begin{minipage}[t]{0.47\textwidth}
\vspace{0pt}
\centering
\begingroup
\setlength{\aboverulesep}{0pt}
\setlength{\belowrulesep}{0pt}
\setlength{\cmidrulesep}{0pt}
\setlength{\minrowclearance}{1.5pt}
\setlength{\tabcolsep}{2.5pt}
\renewcommand{\arraystretch}{1.20}
\definecolor{treeDoc}{HTML}{35664D}
\definecolor{treeCode}{HTML}{005A53}
\definecolor{treeMath}{HTML}{00536B}
\definecolor{treeVL}{HTML}{32487A}
\footnotesize
\resizebox{\linewidth}{!}{%
\begin{tabular}{lcccc}
\toprule
\textbf{Memory} &
\cellcolor{treeDoc!9}\color{treeDoc}\textbf{ParseBench} &
\cellcolor{treeCode!9}\color{treeCode}\textbf{Vision2Code} &
\cellcolor{treeMath!9}\color{treeMath}\textbf{MATH-Vision} &
\cellcolor{treeVL!9}\color{treeVL}\textbf{BabyVision} \\
\midrule
Flat &
\cellcolor{treeDoc!4}52.2 & 
\cellcolor{treeCode!4}57.5 &
\cellcolor{treeMath!4}43.6 &
\cellcolor{treeVL!4}13.2 \\
\textbf{Tree} &
\cellcolor{treeDoc!12}\textbf{67.3} &
\cellcolor{treeCode!12}\textbf{59.9} &
\cellcolor{treeMath!12}\textbf{45.2} &
\cellcolor{treeVL!12}\textbf{15.3} \\
\bottomrule
\end{tabular}%
}
\endgroup

\end{minipage}\hfill
\begin{minipage}[t]{0.50\textwidth}
\vspace{0pt}
\centering
\begingroup
\setlength{\aboverulesep}{0pt}
\setlength{\belowrulesep}{0pt}
\setlength{\cmidrulesep}{0pt}
\setlength{\minrowclearance}{1.5pt}
\setlength{\tabcolsep}{2.5pt}
\renewcommand{\arraystretch}{1.20}
\definecolor{treeDoc}{HTML}{35664D}
\definecolor{treeCode}{HTML}{005A53}
\definecolor{treeMath}{HTML}{00536B}
\definecolor{treeVL}{HTML}{32487A}
\footnotesize
\resizebox{\linewidth}{!}{%
\begin{tabular}{lcccc}
\toprule
\textbf{Memory} &
\cellcolor{treeDoc!9}\color{treeDoc}\textbf{ParseBench} &
\cellcolor{treeCode!9}\color{treeCode}\textbf{Vision2Code} &
\cellcolor{treeMath!9}\color{treeMath}\textbf{MATH-Vision} &
\cellcolor{treeVL!9}\color{treeVL}\textbf{BabyVision} \\
\midrule
No Memory &
\cellcolor{treeDoc!4}63.8 & 
\cellcolor{treeCode!4}65.8 &
\cellcolor{treeMath!4}51.2 &
\cellcolor{treeVL!4}11.1 \\
\textbf{EpiCon} &
\cellcolor{treeDoc!12}\textbf{66.5} &
\cellcolor{treeCode!12}\textbf{67.5} &
\cellcolor{treeMath!12}\textbf{57.2} &
\cellcolor{treeVL!12}\textbf{14.6} \\
\bottomrule
\end{tabular}%
}
\endgroup

\end{minipage}
\vspace{-4mm}
\end{table*}
\textbf{Experience organization.}
Table~\ref{tab:tree_memory_ablation} compares flat and tree memory using lessons derived from the same source questions with the trained MC 2B. The flat bank stores lessons without hierarchical organization or consolidation, while the tree bank organizes and consolidates them for retrieval. Both configurations use the same harness, backbone, attempt budget, and memory-context budget. Tree memory improves all scores, including ParseBench from 52.2 to 67.3 and BabyVision from 13.2 to 15.3. These gains support organizing and consolidating experience for subsequent reuse.

\subsection{Generalization to a Stronger Solver}
\label{sec:exp:stronger_backbones}
We evaluate Codex with GPT-5.6-Luna to test whether EpiCon remains useful with a stronger API-based solver beyond the open-source backbones. The 2B memory components and historical bank remain fixed, without target-specific retraining. Both No Memory and EpiCon make a single solving attempt; EpiCon retrieves experience from the frozen bank. Table~\ref{tab:stronger_backbones} shows improvements on all four benchmarks, including ParseBench from 63.8 to 66.5.

\section{Conclusion}
\label{sec:conclusion}

We presented EpiCon, which connects question-level textual and visual memory co-evolution with the accumulation and reuse of shared experience. Our experiments show that experience can remain useful beyond the system that generated it: another harness can benefit from the existing bank and contribute new experience that improves subsequent solving by the original contributor. These findings support agent collective learning through shared external multimodal memory, allowing systems to build on one another's experience while host model parameters remain fixed.

\bibliography{iclr2027_conference}
\bibliographystyle{iclr2027_conference}
\newpage
\appendix
\section{Data Construction Details}
\label{sec:appendix:data}

\subsection{Source Tasks and Data Collection}
\label{sec:appendix:data:tasks}

Training tasks are drawn from MathNet~\citep{alshammari2026mathnet}, MathV360K~\citep{shi2024math}, ChartQA~\citep{masry2022chartqa}, InfoVQA~\citep{mathew2022infographicvqa}, DocVQA~\citep{mathew2021docvqa}, and ChartNet~\citep{kondic2026chartnet}. They cover three domains: visual mathematics, document understanding, and reasoning over charts and infographics. The source tasks provide the questions and images on which we collect multi-agent system (MAS) executions. Memory updates, lessons, and tree-operation targets are subsequently generated from these execution records.

We collect trajectories using CAMEL~\citep{li2023camel}, Codex~\citep{openai2026codex}, and DeepSeek-Harness~\citep{deepseek-harness2026}. Qwen3.8-27B~\citep{qwen38} generates initial memory updates and operation demonstrations. Qwen3.8-Flash-Next~\citep{qwen3.8flashnext} selectively revises unsuccessful updates, improves lesson quality, and corrects tree-operation targets. The accepted targets provide separate supervision for the Memory Controller and Tree Self-Organizer.

\subsection{Replay-Verified Joint Memory Transformations}
\label{sec:appendix:data:controller}

\paragraph{Joint text--visual targets.}
Given the question, images, previous memory, latest attempt, and correctness feedback, Qwen3.8-27B proposes a complete memory state
\begin{equation}
 m'=(\ell',g',p',b'),
\label{eq:data-joint-target}
\end{equation}
where $\ell'$ is guidance text, $g'$ specifies visual use, and $p'$ and $b'$ identify a source image and normalized bounding box. When $g'=0$, the visual-reference fields are null. The target jointly specifies textual guidance and its associated visual evidence. The evaluator uses reference answers to assess attempts and produce correctness feedback. Qwen3.8-Flash-Next revises updates that fail to support successful solving and refines lessons where needed.

\paragraph{Repair and compression.}
Replay compares the previous and proposed memory states with the same MAS host. Repair transformations are retained when the new memory turns a failed attempt into a successful one. Compress transformations preserve successful solving while shortening text without increasing crop area:
\begin{equation}
\operatorname{tokens}(\ell')<\operatorname{tokens}(\ell),
\qquad
\operatorname{area}(b')\leq\operatorname{area}(b).
\label{eq:data-compression}
\end{equation}
Here $\ell$ and $b$ denote the previous text and visual region; crop area is normalized relative to the source image and is zero when no visual region is retained. The original task images remain available during replay. The final accepted memory, including any teacher revision, is the target selected through replay.

Each training example pairs the update inputs with the accepted memory JSON. The dataset contains approximately 25K memory transformations. Multiple transformations may originate from one source question, so this count denotes supervision examples rather than independent tasks. Replay validates the observed memory transition on the source task, not a guarantee of improvement on every subsequent task.

\subsection{Tree Organization and Retrieval Demonstrations}
\label{sec:appendix:data:tree}

We derive seed lessons from the collected controller memories and organize them into small memory banks. Historical lessons and retrieval queries are separated within each bank. Qwen3.8-27B produces demonstrations of placement, merging, splitting, consolidation, and retrieval; Qwen3.8-Flash-Next selectively revises the operation targets.

The final targets are checked for schema validity, node references, partition coverage, selection constraints, and provenance. Valid empty retrievals are retained when no candidate applies. These checks establish structural validity; tree-operation demonstrations are not individually verified through downstream MAS replay. The dataset contains approximately 6.5K operation demonstrations, which count supervised operations rather than seed lessons or independent questions.

\paragraph{Training and bank construction.}
The supervision datasets train the two memory models. The benchmark-derived construction questions described in Appendix~\ref{sec:appendix:protocol} are subsequently used to build external experience banks with the models fixed. These construction questions add memory content without updating model parameters. Evaluation uses separate questions and frozen historical banks.

\section{Training the Two Memory Policies}
\label{sec:appendix:training}

We independently fine-tune two Qwen3.5-2B models on the datasets in Section~\ref{sec:data}. For a policy $P_\psi$ and its operation dataset $\mathcal D_P$, supervised fine-tuning minimizes the target-sequence loss
\begin{equation}
\mathcal L_P(\psi)=
\mathbb E_{(z,y)\sim\mathcal D_P}
\left[-\sum_{j=1}^{|y|}\log p_\psi(y_j\mid z,y_{<j})\right],
\label{eq:epicon-sft}
\end{equation}
where $z$ contains the operation prompt and its available inputs, and $y$ is the validated memory or operation JSON. Controller targets cover repair and compression of joint text--visual memory. Tree targets cover batch placement, merging, splitting, consolidation, and retrieval; only retrieval examples include images.

Both models use LoRA with rank 16 and scaling factor 32, targeting linear layers while freezing the visual encoder and aligner. Training uses three epochs, a learning rate of $10^{-4}$, maximum sequence length 8,192, per-device batch size 2, and gradient accumulation over 8 steps. Checkpoints are selected by validation loss: step 1,000 for the controller and step 258 for the tree memory self-organizer.

\paragraph{Training and deployment interfaces.}
Offline training-data construction uses correctness feedback to generate and screen memory targets. During evaluation with Codex and DeepSeek-Harness, MC revises memory from the question, images, existing memory, answer, and execution trace. The joint update supervises text, visual use, and the source image region together. Adaptive visual injection controls which stored visual memory enters the next solver input; omitting visual memory from an input does not delete it from the stored state.

\section{Experimental Protocol and Data Splits}
\label{sec:appendix:protocol}

\paragraph{Construction and evaluation sets.}
Table~\ref{tab:memory_evolution_data_splits} lists the benchmark-specific question counts. Experiments that use historical memory share an initial set of 1,000 construction questions: 10 from DocVQA2026, 90 from MP-DocVQA, and 100 from each of the other nine benchmarks. Each memory setting builds its bank from the corresponding MAS execution records, or reuses the specified donor bank in transfer experiments. These questions construct memory content without updating model parameters. No Memory and the question-level ablations do not use a historical bank. The standard evaluation split contains 4,538 questions across eleven benchmarks; experiments reporting a subset of benchmarks use the corresponding evaluation subsets. Construction and evaluation are project-defined splits and are disjoint. For ReasonMap, questions are grouped by their source map before splitting: a map is assigned either to construction or to evaluation, never to both. This map-level separation also applies when adding the second construction set. Construction counts refer to source questions, not the number of accepted memory episodes.

\begin{table}[t]
\centering
\caption{Construction and evaluation questions by benchmark. Initial construction is shared by historical-memory experiments and the first stage of cross-harness memory evolution. Additional construction is used only for memory evolution; its evaluation set excludes both construction sets.}
\label{tab:memory_evolution_data_splits}
\begingroup
\small
\setlength{\tabcolsep}{6pt}
\renewcommand{\arraystretch}{1.1}
\begin{tabular}{@{}lrr@{\hspace{16pt}}rr@{}}
\toprule
\multirow{2}{*}{\textbf{Benchmark}} & \multicolumn{2}{c}{\textbf{Construction}} & \multicolumn{2}{c}{\textbf{Evaluation}} \\
\cmidrule(lr){2-3}\cmidrule(lr){4-5}
& Initial & Additional & Standard & Evolution \\
\midrule
DocVQA2026 & 10 & 0 & 70 & 70 \\
MP-DocVQA & 90 & 100 & 500 & 500 \\
ParseBench & 100 & 50 & 468 & 418 \\
Vision2Code & 100 & 200 & 500 & 500 \\
Omni-I2C & 100 & 150 & 500 & 500 \\
ChartMimic & 100 & 0 & 500 & 500 \\
MATH-Vision & 100 & 100 & 500 & 500 \\
WeMath & 100 & 100 & 500 & 500 \\
WorldBench & 100 & 210 & 500 & 500 \\
ReasonMap & 100 & 40 & 212 & 162 \\
BabyVision & 100 & 50 & 288 & 238 \\
\midrule
\textbf{Total} & \textbf{1,000} & \textbf{1,000} & \textbf{4,538} & \textbf{4,388} \\
\bottomrule
\end{tabular}
\endgroup
\end{table}

\paragraph{Cross-harness memory evolution.}
\label{sec:appendix:roundtrip}
The first construction stage uses the same 1,000 questions as the other historical-memory experiments. A second, non-overlapping set of 1,000 questions is used by the other harness to use and evolve the original bank. Table~\ref{tab:memory_evolution_data_splits} gives the distribution for both stages. The evaluation split contains 4,388 questions and excludes both construction sets. Codex and DeepSeek-Harness exchange the roles of initial construction and subsequent evolution, with Qwen3.8-27B and the two trained 2B memory models fixed. For each evaluated harness, the original and evolved banks are compared on identical questions with one historical retrieval and a single solving attempt, without question-level memory updates. Table~\ref{tab:experience_transfer_roundtrip} follows the same single-attempt protocol as Table~\ref{tab:main_results}; its evaluation split additionally excludes the second construction set.

\paragraph{Evaluation controls.}
Model parameters and historical banks remain frozen during evaluation. The main comparison and all evaluations with historical memory retrieval use a single solving attempt per question, with question-level memory updates disabled. Each attempt includes the harness's standard internal agent and tool interactions. Where enabled in other experiments, question-level updates affect only the current question and are not written back to the bank. Reference answers and official scores are used only for offline evaluation and do not enter memory updates, stopping decisions, or answer selection. We report the last completed answer. Cross-backbone and cross-harness transfer compare one solve with and without the donor bank, with question-level updates disabled. No Memory makes a single solving attempt in every experiment, including the question-level controller comparison, multimodal ablation, and stronger-solver comparison. The question-level controller comparison and multimodal ablation disable historical retrieval and allow up to five solving attempts for memory-enabled settings, including the initial attempt. The system may stop earlier when it judges the current answer ready to return. Their comparisons against No Memory measure memory-guided refinement as a whole, including additional attempts. Memory-enabled settings in the multimodal ablation share the attempt budget. Original task images remain available in the multimodal ablation. Results may be reused across experiments only when the samples, harness, backbone, memory configuration, bank, and inference protocol match; a different evaluation subset requires recomputing scores from the corresponding predictions.

\paragraph{Score and cost reporting.}
We use each benchmark's evaluation metric and give equal weight to each benchmark when computing macro-average scores. Time and token usage are reported separately for MAS execution and memory operations. For each comparison, let $T_0$ and $N_0$ be the corresponding No Memory MAS time and token usage under the same harness, backbone, evaluation set, and baseline protocol. We normalize each cost component $c$ as
\begin{equation}
\widetilde{T}_c=\frac{T_c}{T_0},\qquad
\widetilde{N}_c=\frac{N_c}{N_0},\qquad
c\in\{\mathrm{MAS},\mathrm{Memory}\}.
\label{eq:appendix:cost-normalization}
\end{equation}
Thus, No Memory MAS time and token usage are each $1\times$, and its memory-operation costs are zero. Memory costs use the same MAS denominators rather than a separate memory baseline.

\paragraph{Hardware.}
All experiments are run using NVIDIA H100 GPUs. GPT-5.6-Luna is accessed through its API.

\section{Additional Performance and Cost Comparisons}
\label{sec:appendix:tradeoffs}

\begin{figure*}[htbp]
\centering
\includegraphics[width=\linewidth]{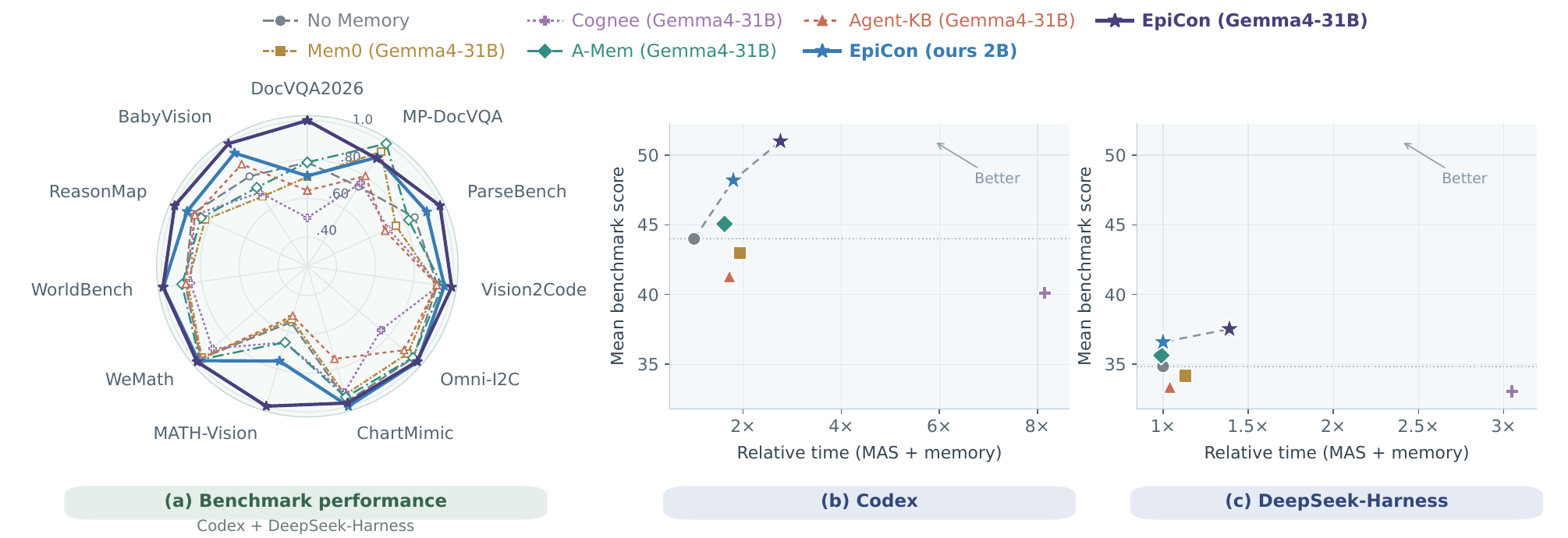}
\caption{Benchmark profiles and performance--time trade-offs with Gemma4-31B, using the results in Table~\ref{tab:main_results}. (a) Scores are averaged across Codex and DeepSeek-Harness, then divided by the best method mean for each benchmark; the radial axis starts at 0.25, as in Figure~\ref{fig:benchmark_cost_maxratio_zoom}. (b) and (c) show performance as the equally weighted mean across eleven benchmarks. Total time sums the normalized MAS and memory costs within each harness. Dashed lines connect No Memory, EpiCon (ours 2B), and EpiCon (Gemma4-31B) as visual guides, not Pareto fronts.}
\label{fig:appendix_gemma_radar_time}
\end{figure*}

\begin{figure*}[htbp]
\centering
\includegraphics[width=0.8\linewidth]{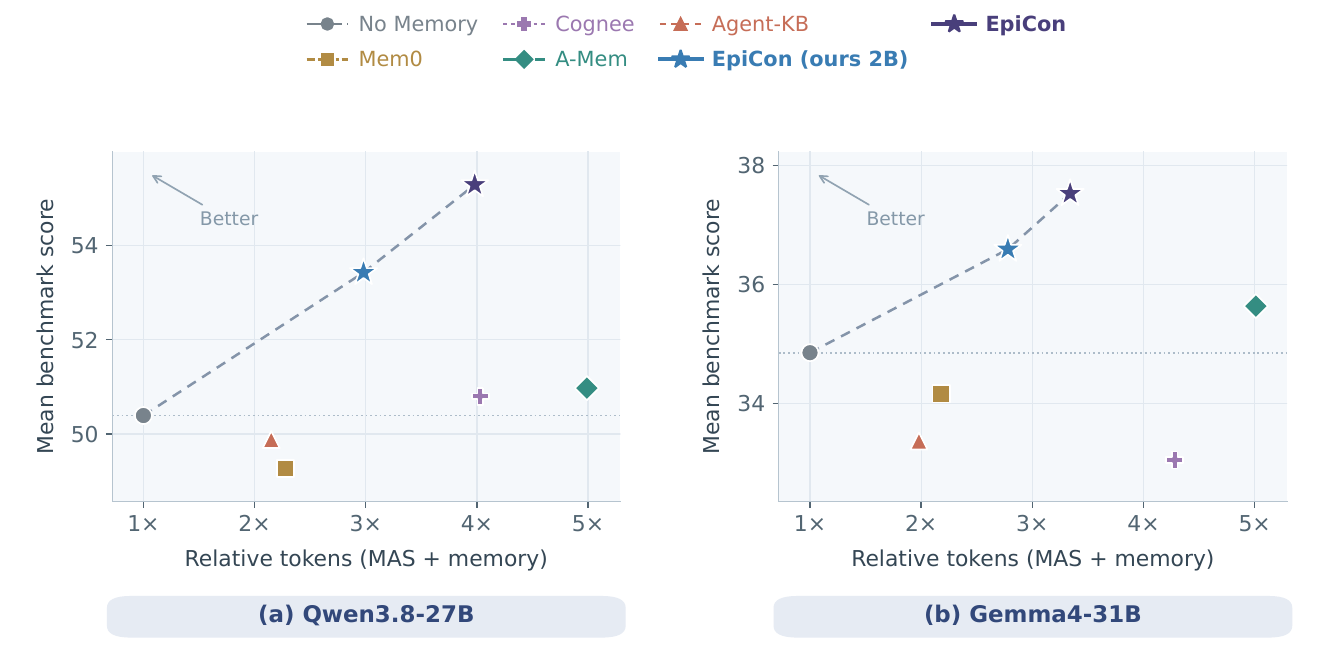}
\caption{Performance--token trade-offs on DeepSeek-Harness with (a) Qwen3.8-27B and (b) Gemma4-31B. Performance is the equally weighted mean across eleven benchmarks in Table~\ref{tab:main_results}. Total token usage sums MAS and memory tokens, normalized by the corresponding No Memory MAS usage for each backbone. Memory methods use the backbone named below each panel, except EpiCon (ours 2B), which uses two independently trained 2B memory models. The panels share the token scale and use separate score ranges. Dashed lines connect No Memory and the two EpiCon variants as visual guides, not Pareto fronts.}
\label{fig:appendix_deepseek_token_tradeoffs}
\end{figure*}

Figures~\ref{fig:appendix_gemma_radar_time}--\ref{fig:appendix_deepseek_token_tradeoffs} extend Figure~\ref{fig:benchmark_cost_maxratio_zoom} using the same main-results measurements. With Gemma4-31B, backbone-sized EpiCon improves the eleven-benchmark mean over No Memory by 7.0 points on Codex and 2.7 points on DeepSeek-Harness; the trained 2B variant gains 4.2 and 1.7 points with lower total time than the backbone-sized variant. Figure~\ref{fig:appendix_deepseek_token_tradeoffs} compares token usage on DeepSeek-Harness with the two backbones. Relative to backbone-sized EpiCon, the trained 2B variant uses 25.1\% fewer total tokens with Qwen3.8-27B and 16.8\% fewer with Gemma4-31B, while yielding lower mean benchmark scores. These results show the performance--token trade-off of the two trained 2B memory models across backbones.


\end{document}